\documentclass[10pt,final,journal]{IEEEtran}
\usepackage{graphicx}
\usepackage{subcaption}
\usepackage{soul}
\usepackage{booktabs}
\usepackage{multirow}
\usepackage{tikz}
\usepackage{comment}
\usepackage{amsmath,amssymb} 

\usepackage{color}
\usepackage[ruled, vlined]{algorithm2e}

\newcommand{\C}{\mathbf{C}}
\newcommand{\D}{\mathbf{D}}

\newcommand{\I}{\mathbf{I}}
\newcommand{\M}{\mathbf{M}}

\newcommand{\X}{\mathbf{X}}

\newcommand{\W}{\mathbf{W}}
\newcommand{\Y}{\mathbf{Y}}

\usepackage[pagebackref=true,breaklinks=true,letterpaper=true,colorlinks,bookmarks=false]{hyperref}

\usepackage[capitalize]{cleveref}
\crefname{section}{Sec.}{Secs.}
\Crefname{section}{Section}{Sections}
\Crefname{table}{Table}{Tables}
\crefname{table}{Tab.}{Tabs.}

\usepackage[accsupp]{axessibility}  

\begin{document}
\title{Autoencoders with Intrinsic Dimension Constraints \\ \hspace{-1cm} for Learning  Low Dimensional Image Representations}
\author{Jianzhang Zheng, Hao Shen, Jian Yang, Xuan Tang, 
\\ Mingsong Chen, Hui Yu, Jielong Guo, Xian Wei$^{\ast}$\\
\thanks{$^\ast$ Corresponding Author\\
Jianzhang Zheng is with the Technical University of Munich, and also with the Fujian Institute of Research on the Structure of Matter, Chinese Academy of Science (e-mail: zhengjianzhang@fjirsm.ac.cn). 
Hao Shen is with the fortiss GmbH (e-mail: shen@fortiss.org).
Jian Yang is with the Information Engineering University (e-mail: jian.yang@tum.de).
Xuan Tang and Mingsong Chen are with the East China Normal University (e-mail: xtang@cee.ecnu.edu.cn; mschen@sei.ecnu.edu.cn).
Hui Yu and Jielong Guo are with the Fujian Institute of Research on the Structure of Matter, Chinese Academy of Science (e-mail:  yuhui@fjirsm.ac.cn; gjl@fjirsm.ac.cn). 
Xian Wei is with the Technical University of Munich (e-mail:xian.wei@tum.de).
}
}

\maketitle
\begin{abstract}
Autoencoders have achieved great success in various computer vision applications.
The autoencoder learns appropriate low dimensional image representations through the self-supervised paradigm, i.e., reconstruction. 
Existing studies mainly focus on the minimizing the reconstruction error on pixel level of image, while ignoring the preservation of Intrinsic Dimension (ID), which is a fundamental geometric property of data representations in Deep Neural Networks (DNNs).
The learning process of DNNs is observed involving highly with the change of the ID of data representations.
Motivated by the important role of ID, in this paper, we propose a novel deep representation learning approach with autoencoder, which incorporates regularization of the global and local ID constraints into the reconstruction of data representations.
This approach not only preserves the global manifold structure of the whole dataset, but also maintains the local manifold structure of the feature maps of each point, which makes the learned low-dimensional features more discriminant and improves the performance of the downstream algorithms. 
To our best knowledge, existing works are rare and limited on exploiting both global and local ID invariant properties on the regularization of autoencoders.
Numerical experimental results on benchmark datasets (Extended Yale B, Caltech101 and ImageNet) show that the resulting regularized learning models achieve better discriminative representations for downstream tasks including image classification and clustering.
\end{abstract}

\section{Introduction}
\begin{figure*}[t!]
\subfloat[]{
\begin{minipage}[c]{0.50\linewidth}
\centering
\includegraphics[width=\linewidth]{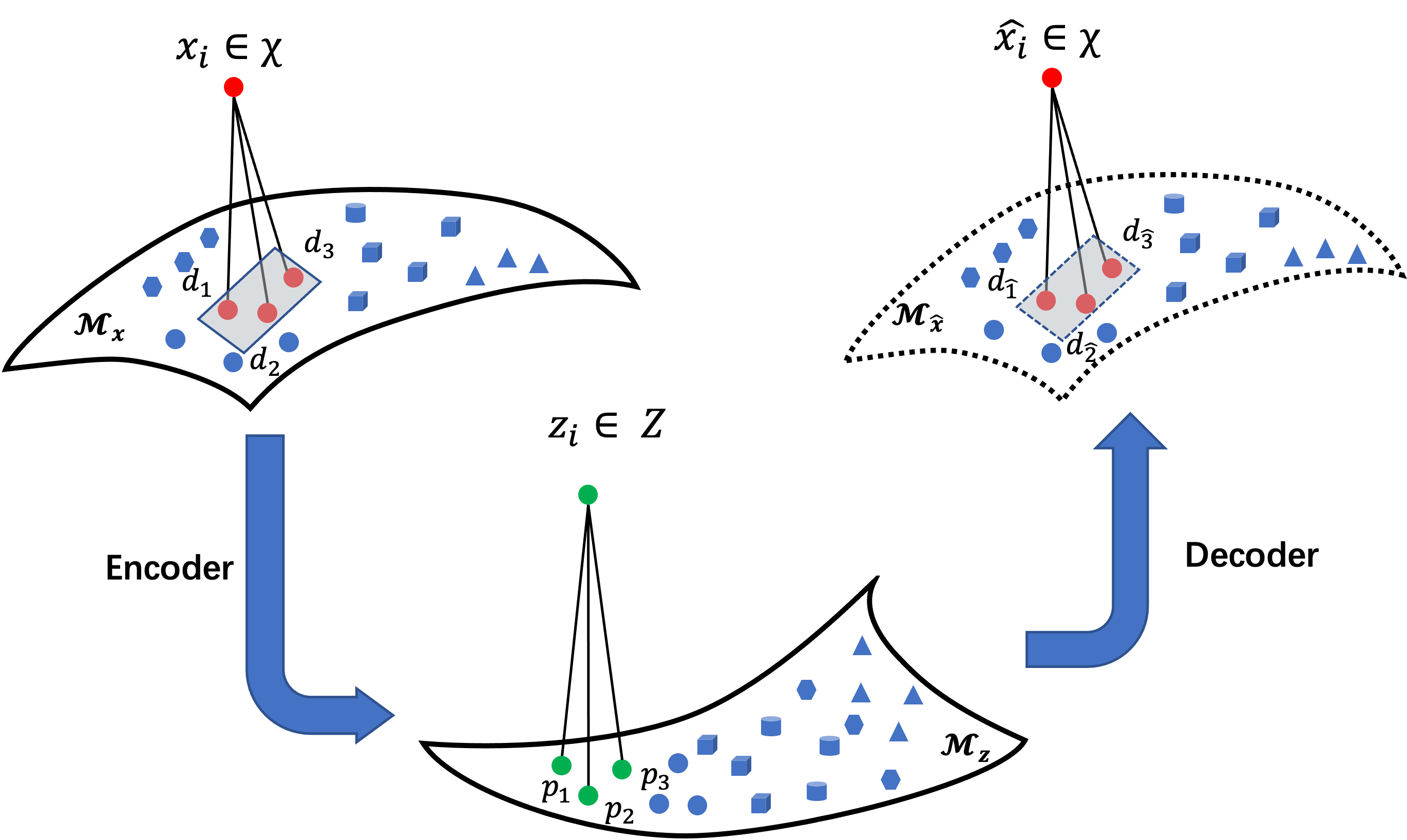}
\end{minipage}
}
\subfloat[]{
\begin{minipage}[c]{0.50\linewidth}
\centering
\includegraphics[width=\linewidth]{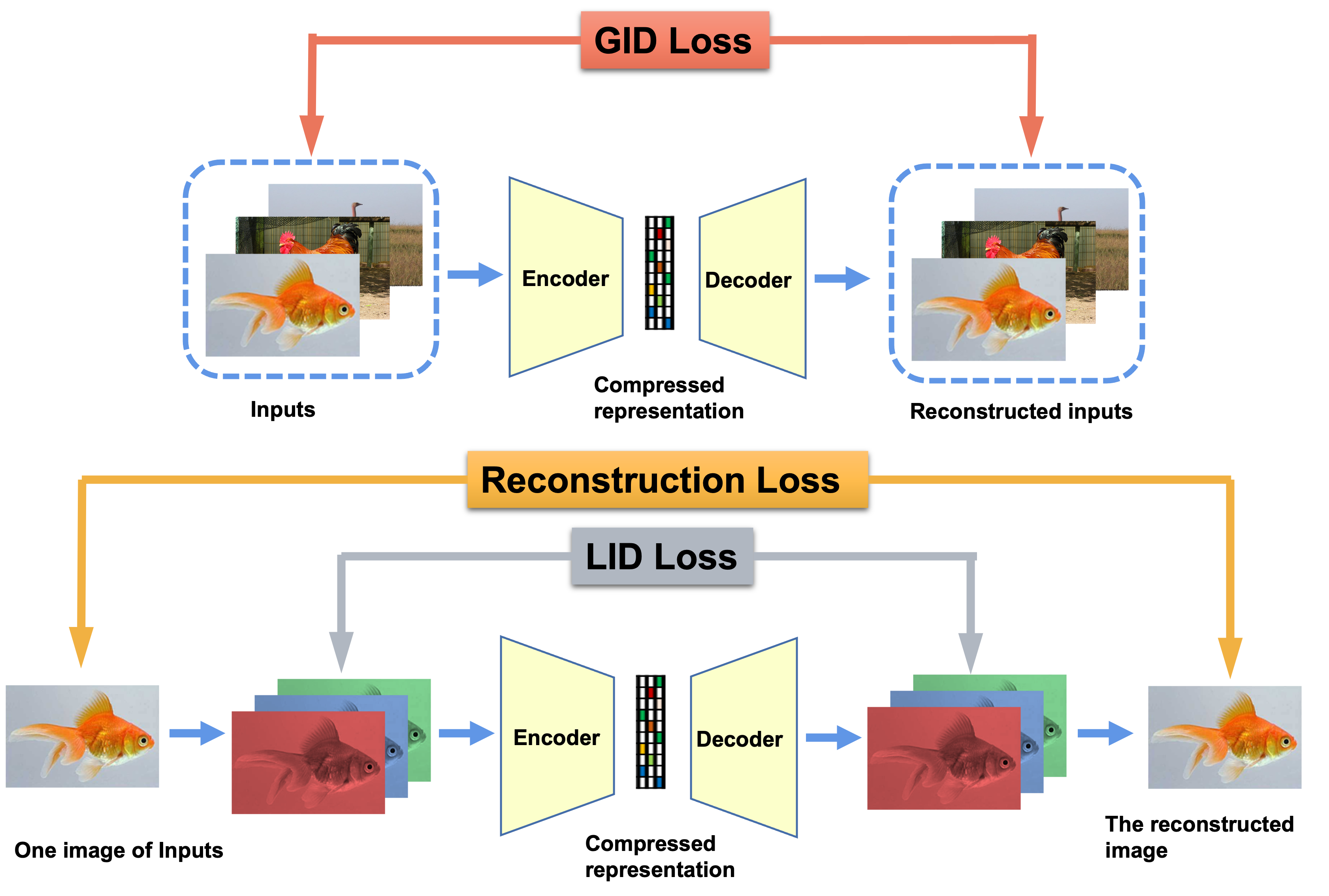}
\end{minipage}
}
\caption{Overview of training paradigm of AE-IDC. GID: global intrinsic dimension; LID: local intrinsic dimension. 
(a) The goal of AE-IDC is to get compressed representations while maintaining data manifold structure globally and locally during the process of encoding and decoding.
(b) An illustration of reconstruction loss, GID loss and LID loss on an image dataset.
}
\label{fig:Overview}
\end{figure*}

Recently, Deep Neural Networks (DNNs) have been successfully applied to various high-dimensional machine learning tasks, 
such as computer vision and natural language processing \cite{chollet2017xception_cvpr, chen2018deeplab_pami, he2017wider_nips}. 
Instead of learning the mapping relationship, DNNs learn the intrinsic geometric structure of data representations and flatten the data in higher layers \cite{Ansuini2019twoNN_nips, brahma2015deep_tnnls}.
The learning process of data representations in DNNs often involves layer-wise changes in features for their inputs, e.g., dimension increase and dimension reduction. 

Intrinsic Dimension (ID)  is a fundamental geometrical property of data representations in DNNs, which is the minimum number of variables or parameters needed to describe points in the ambient space with little information loss \cite{LevinaB04MLE_nips, Facco2017TwoNN_SR,Ansuini2019twoNN_nips,bac2021scikit}.
Correspondingly, the dimension of ambient space is referred to as Extrinsic Dimension (ED). 
Although the ED of data representations is high, data are often concentrated around a low-dimensional manifold \cite{tenenbaum2000global_science, roweis2000nonlinear_science, bengio2013representation_pami}, and the ID of the high-dimensional data representations can be defined as the dimension of the embedded manifold \cite{Ansuini2019twoNN_nips, bac2021scikit}.
The ID can also be defined by comparing with a series of distributions with known dimension, and the dimension of the distribution with the most similar characteristics is the desired ID \cite{Facco2017TwoNN_SR,bac2021scikit}.


As revealed in the research field of self-supervised representation learning \cite{doersch2015unsupervisedICCV, Pathak2016context_cvpr,gidaris2018unsupervised_iclr,He22_MAE_cvpr,bao2022beit_iclr}, the intrinsic structure information of observed data learned by self-supervised models have the capability of discrimination that is useful for downstream tasks.
AutoEncoder (AE) is a classical and effective self-supervised representation learning paradigm through encoding and decoding maps to get compact and discriminative low-dimensional representations \cite{bengio2013representation_pami}.
In recent years, amounts of AE-based models have been proposed and they develop strategies to employ the spatial, temporal, spectral and context structure information of inputs \cite{zhou2019HSI_TGRS, jing2020SelfSupervised_Survey_PAMI, He22_MAE_cvpr}. 
Inspired by the powerful representation capability and success of these strategies, this work explores how the ID can be used to preserve the intrinsic manifold structure of representations improve the performance on the downstream tasks.
To our best of knowledge, it is still an important pending question of how to develop AE that exploit both global and local ID knowledge.

In this paper, we attempt to answer this question by exploiting the ID information in data representations to preserve the global and local manifold structure during transformation. 
Our generic approach can be applied in various forms of AE, but the training paradigm is different from the vanilla AE, which considers global and local ID loss during reconstruction, shown in \cref{fig:Overview}.
The main contributions of this paper are summarized as follows:
\begin{itemize}
\item The proposed method exploits ID information of data representations from global and local perspectives, i.e., the global ID for whole sample data points and the local ID for feature maps from each point.
\item A generic ID-based regularization is incorporated into the current standard autoencoder framework to develop a new training paradigm, coined as AutoEncoder with Intrinsic Dimension Constraint (AE-IDC), which manages to maintain ID invariant along with reconstruction.
\item The AE-IDC achieves more discriminative compressed representations than existing AE models based on Convolutional Neural Networks (CNNs) and Vision Transformers (ViT) on small scale, middle-level scale and large scale datasets:
Extended Yale B \cite{georghiades2001YaleB}, Caltech101 \cite{fei2004Caltech101}, and ImageNet \cite{russakovsky2015ImageNet}.
\end{itemize}

\section{Related work}  
\subsection{Self-supervised Representation Learning} 
Autoencoding is a classical paradigm in self-supervised learning, whhich is widely used in dimensional reduction, denoising and generation.. 
It consists of two parts: the encoder maps the input to latent space, and the decoder reconstructs the input.
There are various self-supervised learners based on autoencoding, such as Principal Component Analysis (PCA) \cite{FlowerBook}, Denoising AE (DAE) \cite{Vincent08DAE}, Variational AE (VAE) \cite{kingma2014VAE}, Generative Adversarial Networks (GAN) \cite{Goodfell14_GAN_NIPS}.
Leveraging modern architectures in DNNs such as Residual Networks (ResNet) \cite{He2016resnet_eccv} and ViT \cite{Dosovitskiy2021_ViT}, it can deal with large-scale data efficiently.
Recently, Masked AutoEncoders (MAE) \cite{He22_MAE_cvpr} and its convolutional variant \cite{gao2022mcmae_nips}, driven by the reconstruction from partially random masked samples, achieve state-of-the-art performances in downstream vision tasks.
The proposed training framework can enhance MAE's learning through imposing ID constraints.

Besides AE-based learners, contrastive learning is another popular paradigm in self-supervised learning.
The contrastive learning methods such as MoCo \cite{he2020MoCo}, SimCLR \cite{chen2020SimCLR} and DINO \cite{caron2021DINO} do not require the model to be able to reconstruct the original input, but instead expects the model to learn discriminative representations in the embedding space by maximizing the distance between different points and minimizing the distance between different augmented representations from the same data point. 

\subsection{Intrinsic Dimension Estimation Techniques} 
The concept of Intrinsic Dimension (ID) has gained significant attention because it helps in learning the underlying manifold of high-dimensional data.
ID is a powerful tool that quantitatively measures the intrinsic geometric structure of data representations.
It also serves as a lower bound of the dimension reduction of the dataset and a measurement of the complexity of the dataset. 
Accurate ID can help to understand the structure of data representations, and guide the design of DNNs in terms of their width and depth.
The overestimation of ID brings additional computational overhead, whereas the underestimation of ID results in significant information loss.

A number of estimation methods have been developed, and they are generally grouped into global or local class \cite{bac2021scikit}.
Global methods consider the entire dataset to provide a single Global Intrinsic Dimension (GID) estimation for the dataset.
Some of the typical algorithms include correlation dimension \cite{Grassberger1983CorrID_TS},  DANCo \cite{ceruti2014danco_PR} and TwoNN \cite{Facco2017TwoNN_SR}.
Local methods analyze each data point's neighborhoods separately, and provide Local Intrinsic Dimension (LID) estimation for each point in the dataset. 
Some of the typical algorithms are Manifold-Adaptive Dimension Estimation (MADE) \cite{farahmand2007manifold_icml}, Maximum Likelihood Estimation (MLE) \cite{LevinaB04MLE_nips}, and Geometry-aware MLE \cite{Gomtsyan2019GeometryAwareML}. 
Both global and local IDs can be repurposed: global ID can be estimated by combining local ID estimations, while local ID can be estimated by applying global ID estimation within a local neighborhood.
In the proposed AE-IDC framework, the ID estimator used is a global class estimator. 

\subsection{The Effect of Intrinsic Dimension on DNNs}
The change of ID is related to the change of geometrical properties of DNNs such as distribution, distance and curvature \cite{Ma2018LID_iclr, Ansuini2019twoNN_nips}, 
so ID is a quantitative characterization for understanding learning behavior from geometrical perspective.
\cite{huang2018mechanisms_PRE} constructed an ID-based framework to understand how compact representations are developed across layers in simplified neural networks.
The work in \cite{nakada2020adaptive_jmlr,latorre2021Effect_nips, Birdal21PersistentHomology_nips} analyzed the effect of the ID on the generalization of DNNs.
Ansuini \textit{et al.} \cite{Ansuini2019twoNN_nips} found that the ID profile in trained DNNs follows a hunchback shape, i.e. the ID first increases and then decreases, and the ID of the last hidden layer is crucial to the classifier's performance.
\cite{Ma18NoisyLabel_icml} found the shift of ID is an indication of the start of overfit in the learning process of DNNs on the dataset with noisy labels.
\cite{jiang2021relationship_CPB} characterized the smoothness of data manifold by the knowledge of data representation's ID. 

Besides the effect on the generalization of DNNs, ID have effect on the robustness of DNNs. 
\cite{Ansuini2019twoNN_nips} attributes the rise of the ID profile to the redundant features, which are irrelevant to final task predictions.
\cite{Pope2021ID_ICLR} revealed that high dimensional datasets are more difficult for DNNs to learn, and the dataset with higher ID value is more vulnerable to be adversarial perturbations.
Moreover, the class with higher ID value is more vulnerable to attacks compared to other classes in the same image dataset.
\cite{Ma2018LID_iclr} found that adversarial attacks can raise the local ID value and train a local ID-based detector to remove adversarial examples from inputs of a classifier.
However, to the best of our knowledge, there are few studies on explicitly controlling GID and LID of data representations to build a self-supervised representation models.

\section{The Proposed Self-supervised Representation Learning Framework with Intrinsic Dimension Regularizations}
\label{method}
\paragraph{Notation.}
Let $\X \in \mathbb{R}^{N \times C\times H \times W}$ be a batch of inputs, where $N, C, H, W$ are batch size, channel number, height and width respectively. 
$\widehat{\X}$ be the reconstruction of $\X$, 
and $m$ is the number of total features. 
$\widetilde{\X}\in \mathbb{R}^{N\times m}$ is reshaped from $\X$, where $m =C\times H \times W$, and each feature of $\widetilde{\X}$ has been centralized for zero-mean.
The covariance matrix of $\X$ is denoted by $\C=\widetilde{\X}^{\top}\widetilde{\X}$.
The data representation after a linear transformation is denoted as $\Y = \widetilde{\X}\W^{\top} \in \mathbb{R}^{N\times n}$, 
where $\W^{\top} \in\mathbb{R}^{m\times n}$ is the transformation matrix.


The synaptic and neural correlation were shown to be capable of controlling the dimensionality of hidden representations in DNNs \cite{huang2018mechanisms_PRE, zhou2021weakly_PRE}.
Based on this phenomenon, Huang \cite{huang2018mechanisms_PRE} defines an ID estimator for the hidden representation at each layer,
\begin{equation}
{\I\D(\X)}=\frac{\left(\operatorname{tr} (\C)\right)^{2}}{\operatorname{tr}\left(\C^2\right)}=\frac{\left(\sum_{i} e_{i}\right)^{2}}{\sum_{i} e_{i}^{2}},
\label{eq:huang_ID}
\end{equation}
where $\C$ is the covariance matrix of data representations, and $e_{i}$ is the eigenvalue of $\C$.
This estimator considers pairwise correlations among synapses, and reveals the mechanism underlying how the synaptic and neural correlations affect dimension reduction \cite{zhou2021weakly_PRE}.
Inspired by its success in characterizing the synaptic and neural correlation in physics and machine learning, in this work, we leverage \cref{eq:huang_ID} to estimate the global and local ID of data representations.
We develop a novel self-supervised training framework to compress the original representations into a discriminative low dimensional space for downstream tasks, which reduces the variation of the global and local ID of representations along with pixel level reconstruction.

\subsection{Control the Global and Local Intrinsic Dimensions of Data Representations.}
Our framework builds upon a reconstruction-based AE,
Stacked AutoEncoders (SAE), which is composed of multiple AEs and has hierarchical architecture. The output of the previous AE is used as the input for the next AE.
The training is two-stage: train single AE in layerwise manner first and then train the all AEs globally. 
Apart from SAE, the framework is also applicable to other AE variants.

The reconstruction loss of AE can be the Mean Squared Error (MSE)  between the reconstructed and original images, 
$$L(\X, \widehat{\X})=\frac{1}{N}\sum_{i=1}^N\left(\widehat{x}_i-x_i\right)^2,$$
or the binary cross entropy loss, 
$$L(\X, \widehat{\X})=-\frac{1}{N}\sum_{i=1}^N\left[x_i \log \left(\widehat{x}_i\right)+\left(1-x_i\right) \log \left(1-\widehat{x}_i\right)\right],$$
where $x_i \in \mathbb{R}^{C\times H \times W}$ is a sample in $\X$.
By minimizing the reconstruction loss, SAE learns a low-dimensional representations.
However, this approach lacks the consideration about the variation of ID after reconstruction. 

To address this issue, this paper proposes an effective framework named AutoEncoder with Intrinsic Dimension Constraint (AE-IDC), which introduce two extra constraints (GID and LID) to regularize the learning of AE. 

The GID describes the geometric structure of the subspaces of varying dimensions from points in the batch. 
To compute the GID, the original inputs (batch, channel, height, weight) are reshaped into (batch, channel$\times$height$\times$weight). The GID is given by
\begin{equation}
{\mathbf{GID}(\X)}=\frac{\left(\operatorname{tr} (\widetilde{\X}^{\top} \widetilde{\X})\right)^{2}}{\operatorname{tr}\left((\widetilde{\X}^{\top} \widetilde{\X})^2\right)},
\end{equation}

\begin{figure}[t!]
  \centering
   \includegraphics[width=\linewidth]{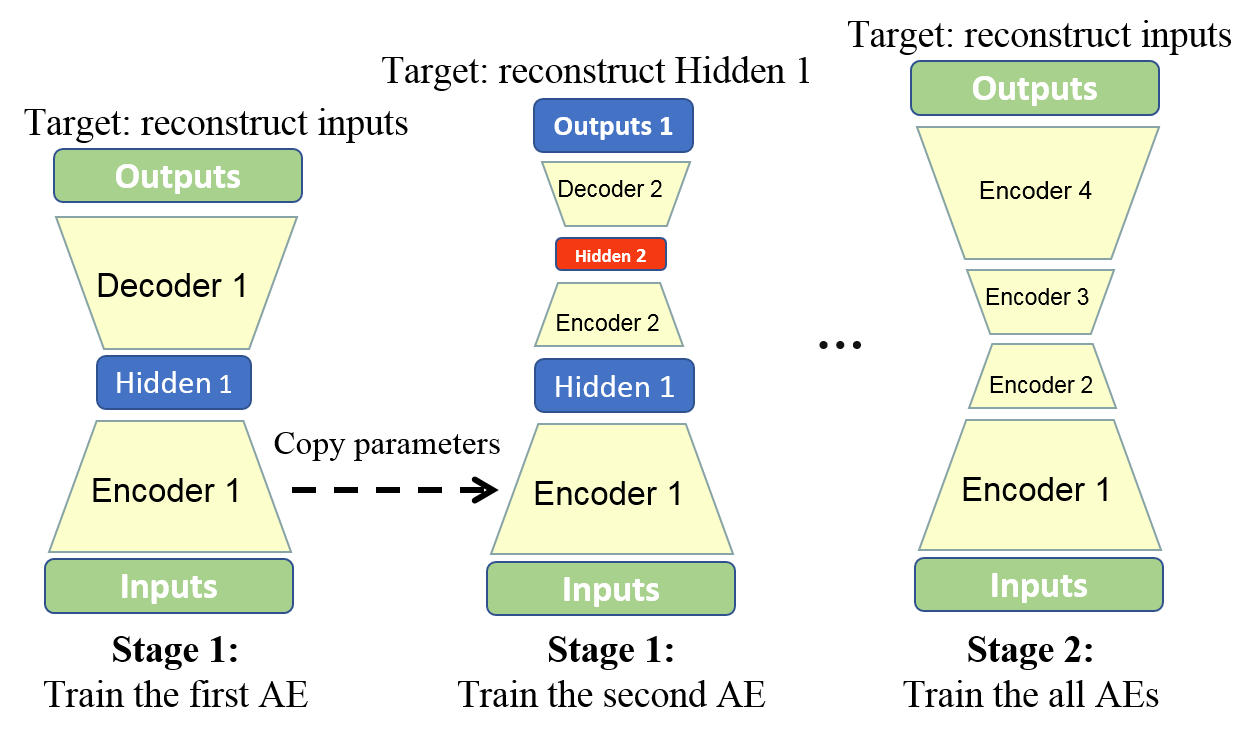}
   \caption{Illustration of Two-stage training for AE-IDC. In this paper, a symmetric AE is used. The first half AEs are undercomplete, whose hidden representations are smaller than inputs; while the second half AEs are overcomplete, whose hidden representations are smaller than inputs.}
   \label{fig:two_stage_train}
\end{figure}

Convolution kernels works as feature extractor and transform an input into multiple similar feature maps. 
We suppose that feature maps of the input indicates the local geometry of it.
Considering that feature maps from the same data point are highly correlated , the space of these feature maps can be used as the description of the local geometric structure for the point.
We regard the ID of such space as the LID of the point. 
To compute the LID, the original sample (channel, height, weight) is reshaped into (channel, height$\times$weight) first. The LID is then given by 
\begin{equation}
{\mathbf{LID}(x_i)}=\frac{\left(\operatorname{tr}(\M^{\top} \M)\right)^{2}}{\operatorname{tr}\left((\mathbf{\M^{\top} \M}\right)^2)}, where \, \M\in \mathbb{R}^{C\times HW}.
\end{equation}

It is intuitive and reasonable that the variation of GID and LID between the original inputs and reconstruction needs to be as small as possible.
Motivated by this consideration, we incorporate these two ID constraints into the reconstruction loss function to encourage the SAE to maintain ID.  
Formally, the objective optimized by a AE-IDC is 
\begin{equation}
\begin{aligned}
\mathcal{J}_{\mathrm{AE-IDC}} = & L(\X, \widehat{\X})+\lambda_1 (\mathrm{GID}(\X)-\mathrm{GID}(\widehat{\X}))^2\\
& +\lambda_2\sum_{i = 1}^{N} (\mathrm{LID}(x_i)-\mathrm{LID}(\widehat{x}_i))^2, 
\end{aligned}
\label{eq:total_Loss}
\end{equation}
where $\lambda_1$ and $\lambda_2$ are hyper-parameters controlling the strength of the corresponding regularization. \cref{fig:vis_32d} depict the performances of AE-IDC with respect to different weighing factors $\lambda_1$ and $\lambda_2$ in the downstream classification task.

Given most loss functions of the DNN models are mainly optimized by the backpropagation \cite{FlowerBook} for a batch of samples from the dataset, the differentiable geometrical structure characterization within the batch is needed.
The reconstruction loss has been proven to be differentiable. The differential of IDC with respect to weight $W$ of networks is computed as
\begin{equation}
\begin{aligned}
\frac{\partial}{\partial \W}&\left(\frac{\operatorname{tr}\left(\W  \X^{\top}   \X \W^{\top} \right)^2}{\operatorname{tr}\left(\W  \X^{\top}   \X \W^{\top}\W  \X^{\top}   \X \W^{\top}\right)} \right)= \\
&\left(4  t_2\right) / t_4 \cdot T_0-\left(4  t_2^2\right) / t_4^2  \cdot T_3,\\
where\,&\hspace{2cm}T_0=\W \X^{\top} \X \\
&\hspace{2cm}T_1=T_0 \W^{\top} \\
&\hspace{2cm}t_2=\operatorname{tr}\left(T_1\right) \\
&\hspace{2cm}T_3=\left(T_1 \W \X^{\top} \X\right) \\
&\hspace{2cm}t_4=\operatorname{tr}\left(T_3  \W^{\top} \right).
\end{aligned}
\end{equation}
Therefore, \cref{eq:total_Loss} is differentiable and can be incorporated into the back propagation.

\subsection{Two-stage Training of Autoencoder with Intrinsic Dimension Constraints}

\begin{algorithm}[!t]
    \caption{Training of AE-IDC.}
    \label{algorithm:Training of AE-IDC}
    \LinesNumbered
    \KwIn {$\mathcal{X}$: a dataset of clean examples.\\
$\{\mathrm{AE}_j\}_{j=1}^{L}$: $L$ AEs.\\
$f_{enc}^i$: the encoder of $\mathrm{AE}_i$;
$f_{dec}^i$: the decoder of $\mathrm{AE}_i$.\\
}
    \KwOut {The encoder part of the SAE.} 
    $\#$ Layerwise training for each AE.\\ 
    \For{i = 1 to l }{
        Freeze all the previous AEs $\{\mathrm{AE}_j\}_{j=1}^{i-1}$.\\
        \For{sample a batch $\X$ from $\mathcal{X}$}{
            input of $\mathrm{AE}_i$: $\X_i=f_{enc}^{i-1}(\X_{i-1})$, where $\X_1=\X$. \\
            compute $loss$ according to \cref{eq:total_Loss}.\\
            perform BP to update parameters of  $\mathrm{AE}_i$ by minimizing $loss$.
        }
        
    }
    $\#$ Global training for whole AEs.\\
    Stack all the encoders $\{f_{enc}^i\}_{i=1}^{L}$ to construct a SAE.\\
    \For{sample a batch $\X$ from $\mathcal{X}$}{
        compute $loss$ according to \cref{eq:total_Loss}.\\
        perform BP to update parameters of SAE by minimizing $loss$.
        }

\end{algorithm}

In the framework, we use an L-layer symmetric SAE, where the first $L/2$ layers perform encoding and the second $L/2$ layers perform decoding, shown in \cref{fig:two_stage_train}.
Note that each layer is a separate AE, consisting of encoder and decoder parts. 
The first half AEs are undercomplete, and the second half AEs are overcomplete.

The training framework of AE-IDC is summarized in \cref{algorithm:Training of AE-IDC}.
The framework also follows a two-stage paradigm.
In the first stage, train the group of AEs in layerwise manner. During training the $i^{th}$ AE, compute the reconstrution, and GID and LID loss according to this AE's input and output, then update the parameters of $\mathrm{AE}_i$ locally.
In the second stage, stack all the pretrained AEs' encoder and perform end-to-end training to update the parameters of $\{f_{enc}^i\}_{i=1}^{L}$ globally.
At inference, we only take the first half of the trained AE-IDC to perform feature extraction for downstream tasks.

\section{Experiments}
To validate the proposed AE-IDC, we first investigate the effect of IDC imposed on the training of models.
Then, evaluate the feature extracting performance of the proposed AE-IDC on two downstream tasks: image classification and clustering.
Finally, conduct extensive ablation studies to analyze the impact of different components of AE-IDC.

\paragraph{Datasets.}
In this section, the performance of the proposed method AE-IDC is validated by experiments on three benchmark image datasets including Extended Yale B, Caltech101, and ImageNet. 
ImageNet10 is a subset of ImageNet-1K, which consists of ten classes selected from ImageNet-1K.
It provides a fast test tool on on ImageNet without loss of generality. 
The configurations about the splitting of datasets in this work are summarized in \cref{tab:Dataset}.

\begin{table}[!t]
\begin{center}
\begin{sc}
\begin{small}
\begin{tabular}{| c@{\hspace{5pt}} |c @{\hspace{2pt}}| c| c|}
\hline
{\bf Dataset} & {\bf Training Set}  & {\bf Testing Set}& {\bf Class}\\
\hline
\hline
{\bf Extended Yale B \cite{georghiades2001YaleB}} & 2314 & 1874 &38 \\
\hline
{\bf Caltech101 \cite{fei2004Caltech101}}  & 6907 & 1770 &101 \\
\hline
{\bf ImageNet10 \cite{russakovsky2015ImageNet}}   & 13000 & 478 &10 \\
\hline
{\bf ImageNet-1K \cite{russakovsky2015ImageNet}}   & 1281167 & 50000 &1000 \\
\hline
\end{tabular}
\end{small}
\end{sc}
\end{center}
\caption{Configuration of Datasets}
\label{tab:Dataset}
\end{table}

\paragraph{Implementation Settings.}
We mainly use convolution, deconvolution, maxpooling and upsampling layers to construct SAE.
The details about architectures of AE used in the following subsections can refer to appendix.
For large-scale dataset, we adopt the standard ViT-Base (ViT-B) \cite{Dosovitskiy2021_ViT}.
Weights of regularizers are set $\lambda_1=0.1$ and $\lambda_2=0.1$ as default.
All experiments in this paper are conducted on an NVIDIA RTX 3090 GPU (2 GB memory).
And the codes for the reproduction of our work will be available at Github.

\subsection{Analysis of AE-IDC's Learning Process}

\begin{figure*}[t!]
  \centering
   \includegraphics[width=\linewidth]{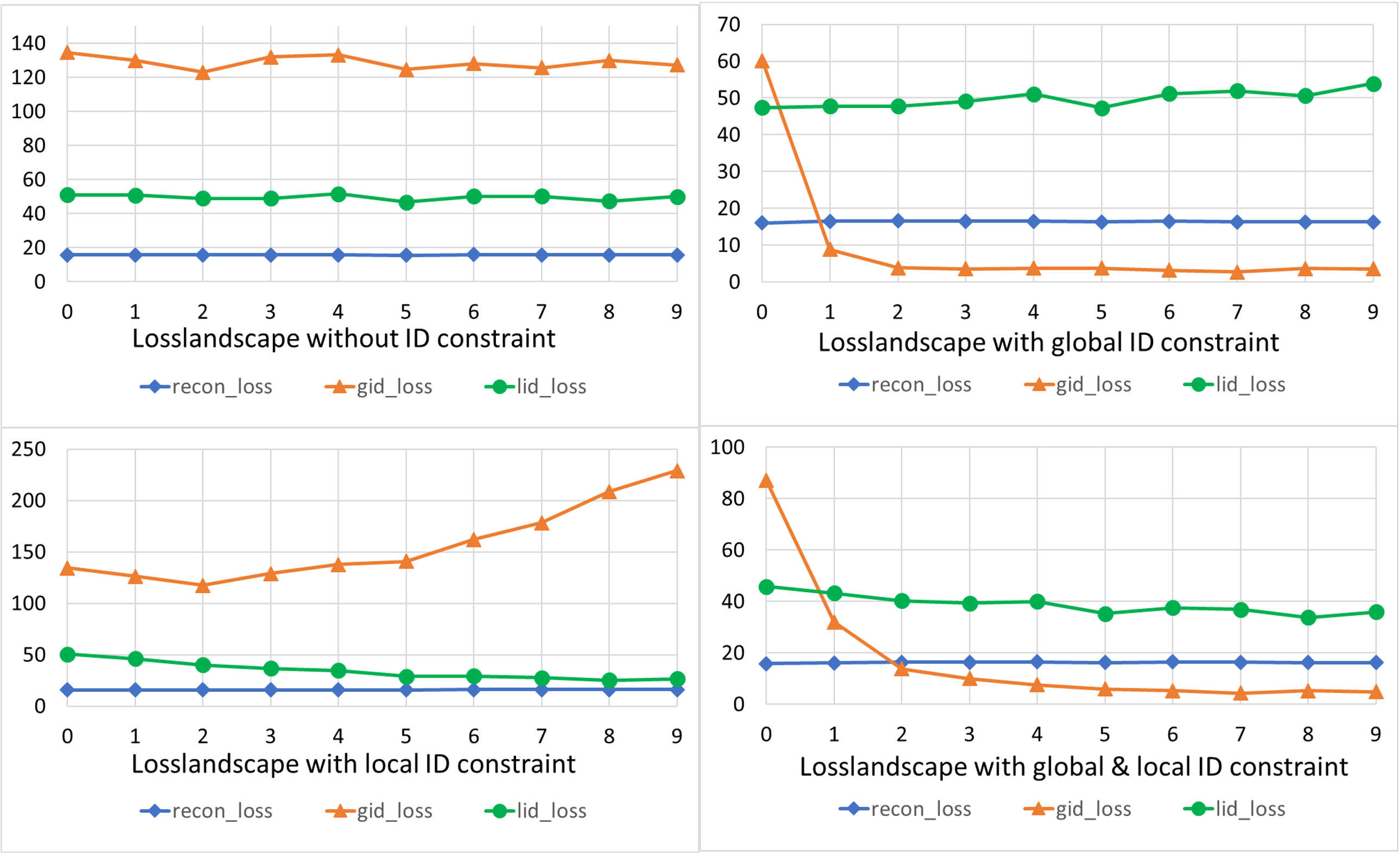}
   \caption{Loss landscapes of MAE and its variants with different intrinsic dimension constraints on ImageNet10. The horizontal axis represents the number of epochs, and the vertical axis represents the loss value. 
   All these four MAEs are initialized with the same pretrained model.}
   \label{fig:vis_LossProfile}
\end{figure*}

\paragraph{CNNs-based AE-IDC.}
\begin{figure}[t!]
  \centering
   \includegraphics[width=\linewidth]{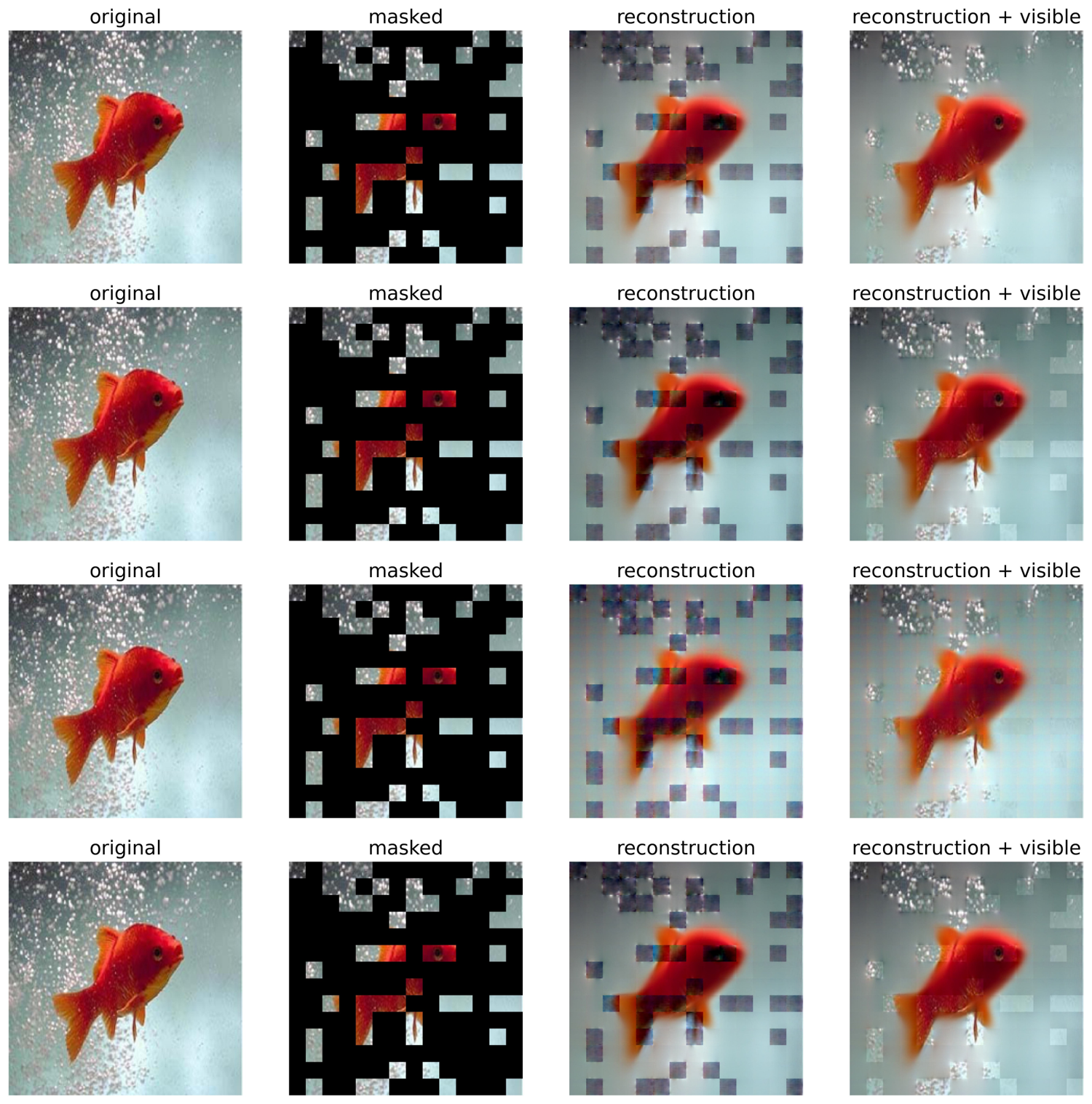}
   \caption{
   Reconstructions of a 75\% masked image from ImageNet through MAE, MAE with GID regularizer, MAE with a LID regularizer and MAE-IDC, from top to bottom. }
   \label{fig:vis_recon}
\end{figure}

To simplify the analysis, we analyze the CNNs model used in Extended Yale B, which only consists of four convolutional layers, where the extrinsic dimension of data representations is (3,32,32)-(12,16,16)-(24,8,8)-(12,16,16)-(3,32,32) from inputs to reconstructions.
The figure depicting the full loss landscapes of two SAE models with and without IDC during two-stage training is attached in appendix.
The reconstruction loss of both models gradually converges at the end of 100 epochs' training in the layerwise and global training stages. 
But SAE-IDC's reconstruction loss is a bit higher than vanilla SAE. This is because the SAE-IDC is regularized to learn a more abstract embedding feature space instead of a simple reconstruction of pixels.
For global and local ID loss, SAE-IDC achieves smoother training curves and quicker convergences in the layerwise and global training stages, compared to vanilla SAE. It can obviously observed that for LID loss vanilla SAE do not converge at the first step layerwise training and the final global training. 
We suppose that the convergence of reconstruction, global and local ID loss is the reason for SAE-IDC to extract discriminative representations for downstream tasks.

\begin{figure}[h!]
  \centering
    \begin{subfigure}{0.47\textwidth}
      \includegraphics[width=\linewidth]{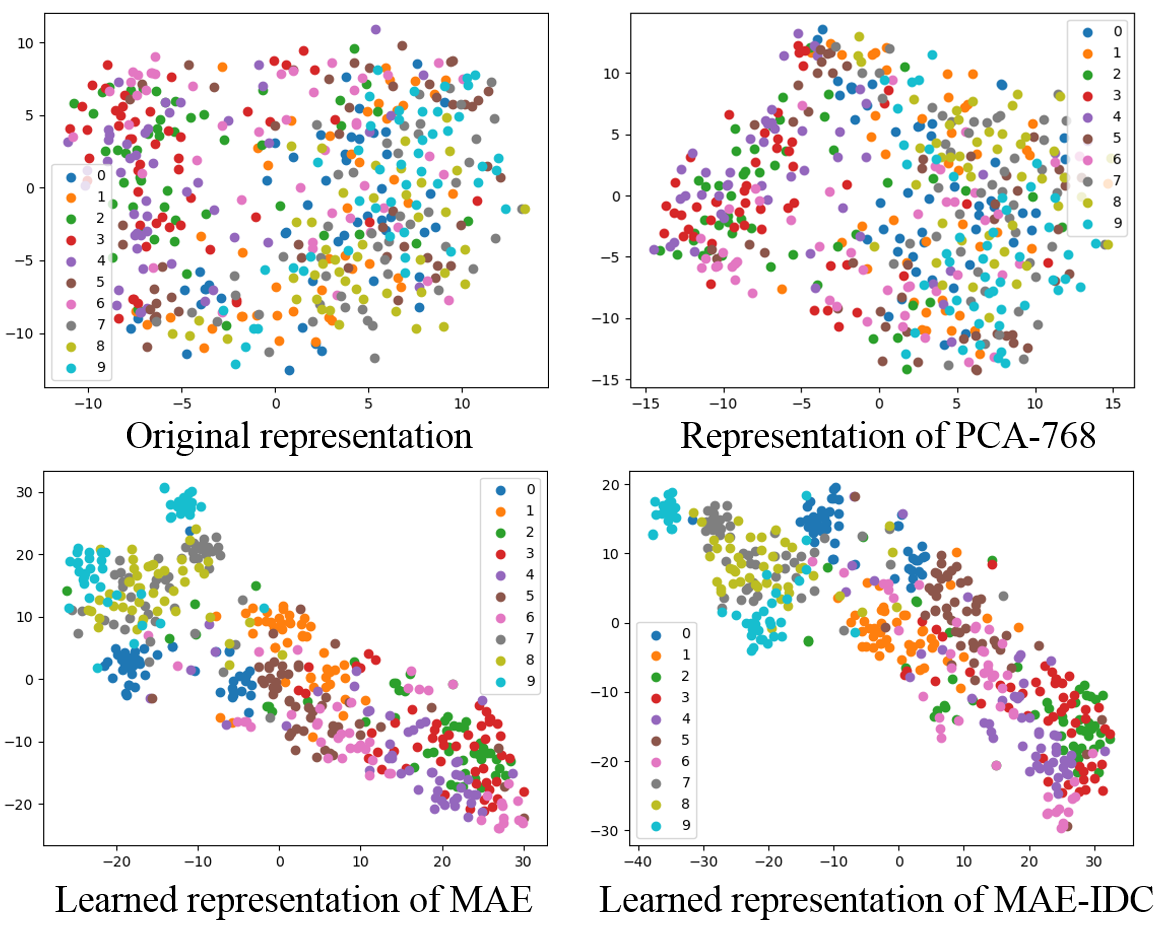}
        \label{fig:sub1}
    \end{subfigure}   
    \begin{subfigure}{0.47\textwidth}
      \includegraphics[width=\linewidth]{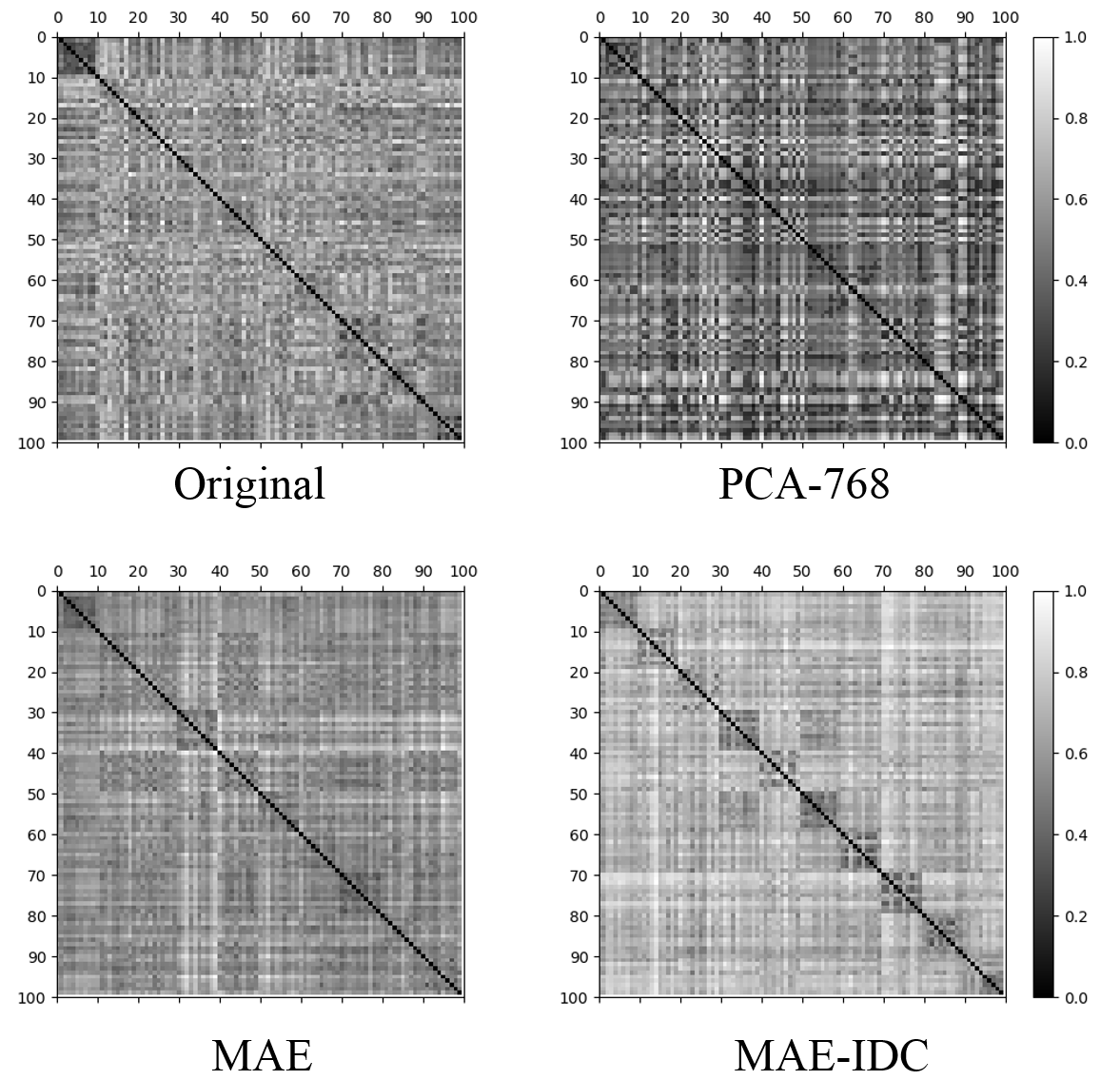}
        \label{fig:sub2}
    \end{subfigure}
\caption{ Visualization of self-supervisedly learned representations on the test set of ImageNet10 using t-SNE (Top group),
and geodesic distance distributions between ten classes from ImageNet10 with each class has ten images (Bottom group).
The geodesic distance is computed approximately by KNN graph euclidean distance where K=15, and is linearly rescaled to [0, 1] for the convenient of visualization. 
The distribution of geodesic distance is decentralized in the original space and the linear subspace by PCA.
Compared to MAE, MAE-IDC is more centralized in diagonal, which means it keeps the manifold structure, i.e., the similarity in the same class and dissimilarity between different classes.}
\label{fig:vis_geodesic_distance}
\end{figure}

\paragraph{ViT-based AE-IDC.}
MAE \cite{He22_MAE_cvpr}, a form of DAE, is a State-Of-The-Art (SOTA) self-supervised learner in AE-based learners.
However, MAE only utilizes MSE as its reconstruction target, which ignore the geometric structure information. 
We will show in following subsection that the enhanced MAE under the proposed IDC, dubbed as MAE-IDC, will unleash the potential of MAE.
\cref{fig:vis_LossProfile} shows the landscape of three kinds of loss. We use pretrained model from \cite{He22_MAE_cvpr} to initalize all models.
The MAE without ID constraint has reached its optimal point at the start, resulting in its losses remaining nearly constant. Conversely, the MAE variants with ID constraints converge quickly after ten epochs, with the GID loss and LID loss dropping after imposing the corresponding regularizer.
%
All these MAE variants keep the reconstruction ability, shown in \cref{fig:vis_recon}.

\subsection{Evaluation of Representations on Downstream Tasks}
We choose image classification and clustering as downstream tasks.
The performance of embedded representations is evaluated on the classification task using K-Nearest Neighbor (KNN) algorithm, and on the clustering task using K-means algorithm. 
The KNN and K-means algorithms provide a fast test, without the need to carry on a heavy end-to-end fine-tuning, and also provide relative fairness for comparison. 
The number of time to run k-means is ten.

\paragraph{Results on Classification Tasks.}
To demonstrate the generality of the proposed algorithmic framework, we apply this framework into two other widely used AE variants, i.e., DAE and sparse AE. %
As seen in \cref{tab:KNN}, for CNNs-based models, AE-IDC outperforms AE without IDC on all three datasets by $1\%\sim5\%$.
We also compare AE-IDC with SOTA self-supervised learning methods on ImageNet10 and ImageNet-1K. 
For fair comparison, let the compared methods use its public official fine-tuned models without any modification. 
as shown in Table~\ref{tab:KNN}. 
Though MAE-IDC only wins the best performance by $0.1\%$ on ImageNet-1K, its training process is more concise and easy to understand.

\begin{table}[!t]
\begin{small}
\begin{tabular}{|cc@{\hspace{4pt}}cccc|}
\hline
\textbf{Method} & \textbf{Arch.} & \textbf{Dim.} & \textbf{k=5}   & \textbf{k=10}  & \textbf{k=15}  \\ \hline
\hline
\multicolumn{6}{|c|}{Extended Yale B}        \\ \hline
SAE     & CNNs      &(24,8,8)     & 82.28      & 81.75      &   79.19    \\ 
SAE-IDC     & CNNs  &(24,8,8)     &\underline{\textbf{86.50}}       & \underline{\textbf{83.78}}      &   \underline{\textbf{81.96}}    \\ 
DAE     & CNNs      &(24,8,8)     & 66.70      & 62.75      &   59.34    \\ 
DAE-IDC     & CNNs  &(24,8,8)     &\underline{\textbf{69.65}}       & \underline{\textbf{65.47}}      &   \underline{\textbf{62.65}}    \\\
SparseAE     & CNNs      &(24,8,8)     & 82.07      & 81.22      &   81.22   \\ 
SparseAE-IDC     & CNNs  &(24,8,8)     &\underline{\textbf{84.85}}       & \underline{\textbf{84.63}}      &   \underline{\textbf{83.30}}    \\\hline
\multicolumn{6}{|c|}{Caltech101}             \\ \hline
SAE     & CNNs      & (24,28,28)    & 49.49      & 43.39      & 39.66      \\ 
SAE-IDC     & CNNs  &(24,28,28)     & \underline{\textbf{50.33}}      &\underline{\textbf{45.48}}       &\underline{\textbf{42.20}}       \\\hline 
\multicolumn{6}{|c|}{ImageNet10}             \\ \hline
SAE     & CNNs      & (512,7,7)    & 34.94      & 33.47      & 33.26      \\ 
SAE-IDC     & CNNs  &(512,7,7)     & \underline{\textbf{35.77}}      &\underline{\textbf{36.19}}       &\underline{\textbf{35.56}}       \\
MAE\cite{He22_MAE_cvpr}  & ViT-B & 768 & 74.89 & 75.52 & 75.94 \\
MAE-IDC         & ViT-B             & 768          & \underline{\textbf{75.94}} & \underline{\textbf{76.56}} & \underline{\textbf{76.98}} \\
MoCo v3 \cite{mocoV3_21_iccv} & ViT-B & 768 & 27.61 & 28.03 & 29.91 \\
DINO \cite{caron2021DINO} & ViT-B & 768 & 73.22 & 73.64 & 76.35 \\ \hline
\multicolumn{6}{|c|}{ImageNet-1K}             \\ \hline
MAE\cite{He22_MAE_cvpr}  & ViT-B & 768 & 49.03 & 45.82 & 43.94 \\
MAE-IDC         & ViT-B             & 768          & \underline{49.14} & \underline{45.93} & \underline{44.06} \\
MoCo v3 \cite{mocoV3_21_iccv} & ViT-B & 768 & 27.61 & 28.03 & 29.91 \\
DINO \cite{caron2021DINO} & ViT-B & 768 & 67.32 & 63.97 & 62.30 \\ \hline
%
\end{tabular}
\end{small}
\caption{
Classification performance (metric: Average Top-1 Accuracy(\%)) on Extended Yale B, Caltech-101 and ImageNet datasets, where a KNN classifier is applied after feature extraction. } 
\label{tab:KNN}
\end{table}

\paragraph{Results on Clustering Tasks.} 
The metrics for evaluation are Adjusted Mutual Index (AMI) and Adjusted Rand Index (ARI). 
All models are pretrained on ImageNet-1K.
The results in \cref{tab:K-means} shows MAE-IDC's advantage over MAE on clustering task. This is in coordinated with visualization results in \cref{fig:vis_geodesic_distance}.
Although MAE-IDC falls behind the SOTA contrastive learning DINO, the gap shrinks on Caltech101.

\begin{table}[!ht]
\begin{center}
\begin{small}
\begin{tabular}{|c|c|c|}
\hline
\textbf{Method} & \textbf{ImageNet10}   & \textbf{Caltech101}   \\ \hline\hline
MAE\cite{He22_MAE_cvpr}   & 0.280$\|$0.451 & 0.293$\|$0.532  \\
MAE-IDC                              & \textbf{ 0.291$\|$0.467} & \textbf{0.313$\|$0.542} \\ \hline
MoCo v3 \cite{mocoV3_21_iccv}  & 0.090$\|$0.189 & 0.047$\|$0.112 \\
DINO \cite{caron2021DINO}  & 0.491$\|$0.645 & 0.328$\|$0.576\\ \hline
\end{tabular}
\end{small}
\end{center}
\caption{
Clustering performance (metric: ARI$\|$AMI) of the proposed AE-IDC with comparisons to three SOTA self-supervised methods, where K-means algorithm applied after feature extraction. The configuration of architecture and embedding dimension is same with \cref{tab:KNN}.} 
\label{tab:K-means}
\end{table}

\subsection{Ablation studies}
In this subsection, we ablate the design of AE-IDC, and analyze the impacts of elements in loss function, stagewise training and weights of regularizers for the performance of AE-IDC.

\paragraph{Elements in Loss Function.}
We demonstrate the classification accuracy of models trained with three variants of loss Function: $\textbf{Reconstruction+GID}$,  $\textbf{Reconstruction+LID}$ and  $\textbf{GID+LID}$.
As shown in \cref{tab:Abl_LossItem}, the reconstruction loss is the most significant factor influencing the quality of the learned representations. While using GID and LID loss separately is insufficient for the learning of AE-IDC. Therefore, ID constraints should be combined with the reconstruction loss, and there exists a synergistic relationship between the GID regularizer and LID regularizer.

\begin{table}[t!]
\begin{center}
\begin{sc}
\begin{small}
\begin{tabular}{|c|c|}
\hline
{\bf Loss Item} & {\bf KNN Accuracy} \\
\hline
\hline
{\bf Reconstruction (Baseline)} & 82.28  \\
\hline
{\bf Reconstruction  + GID}  & 83.99  \\
\hline
{\bf Reconstruction  + LID} & 83.78  \\
\hline
{\bf Reconstruction  + GID  + LID} & {\bf 86.50}  \\
\hline
{\bf GID  + LID } & 59.98  \\
\hline
\end{tabular}
\end{small}
\end{sc}
\end{center}
\caption{
Impact of the each component in the proposed loss function (\cref{eq:total_Loss}) to the KNN recognition rate on the Extended Yale B.}
\label{tab:Abl_LossItem}
\end{table}

\paragraph{Stagewise Training.}
Here we compare the two-stage training with two training variants: one-stage layerwise training and one-stage global learning. 
\cref{tab:Abl_TrainParadigm} demonstrate the efficacy of the two-stage training, which is superior than both one-stage layerwise training and one-stage global learning. 
The results in \cref{tab:Abl_TrainParadigm} also validate that one-stage layerwise training or global training can learn more effective representations, compared to the baseline. 
There in the resource-limited situations, using one-stage global training can be an option to reduce the training time.

\begin{table}[!t]
\begin{center}
\begin{sc}
\begin{small}
\begin{tabular}{|c|c|}
\hline
{\bf Training Paradigm} & {\bf KNN Accuracy} \\
\hline
\hline
{\bf Baseline } & 82.28  \\
\hline
{\bf Layerwise training } & 83.88  \\
\hline
{\bf Global training } & 85.92  \\
\hline
{\bf Layerwise training  + Global training} & {\bf 86.50}  \\
\hline
\end{tabular}
\end{small}
\end{sc}
\end{center}
\caption{
Impact of each stage in the proposed two-stage training paradigm of AE-IDC in \cref{algorithm:Training of AE-IDC} to the KNN recognition rate on the Extended Yale B. 
}
\label{tab:Abl_TrainParadigm}
\end{table}

\begin{figure}[t!]
  \centering
   \includegraphics[width=\linewidth]{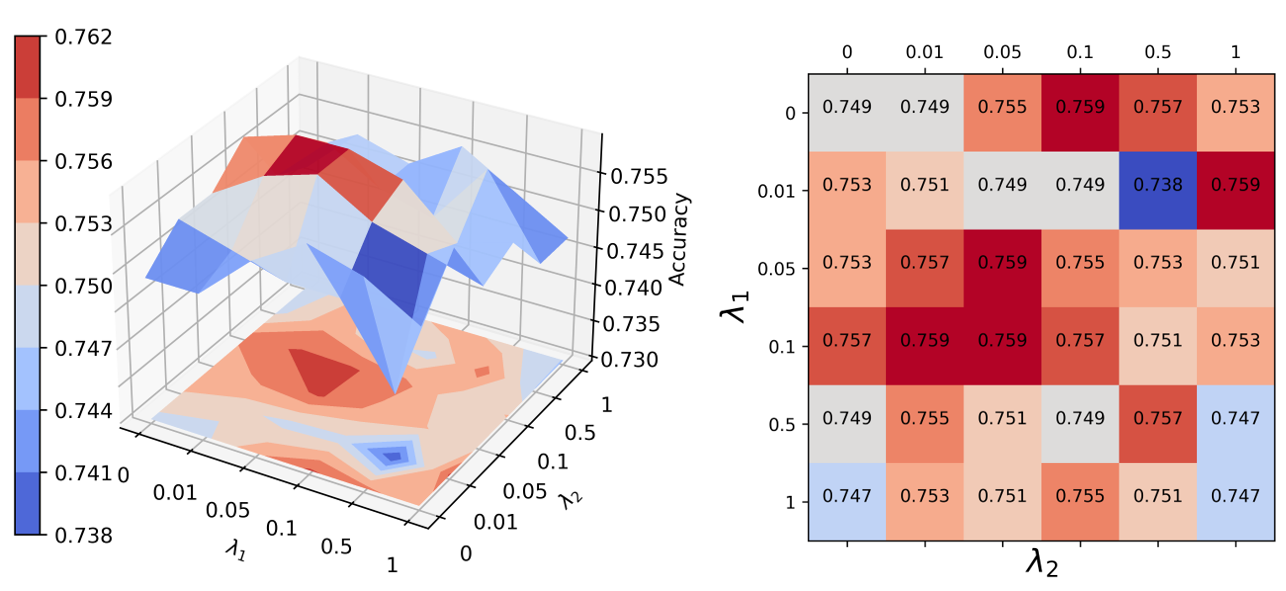}
\caption{
Impact of the weights of regularizers to the recognition rate of KNN on ImageNet10. 
The 3D contour plot is generated from the discrete heatmaps displayed on the right-hand side.}
   \label{fig:vis_32d}
\end{figure}

\paragraph{Weight of Regularizers.}
We investigate the impact of two critical hyper-parameters $\lambda_1$ and $\lambda_2$: the weights for the GID regularizer and LID regularizer respectively. \cref{fig:vis_32d} demonstrate that IDC is not highly sensitive to the choice of these weights. This alleviates the need for extensive fine-tuning and facilitates the implementation of our approach on the customized dataset.

\section{Conclusions}
In this work, we proposed a novel regularized autoencoder for representation learning, which exploits data representations' global and local ID information, coined as AE-IDC.
Specifically, we regard the ID of the manifold formed by the same batch of images as the estimation of the global ID of this batch of images, and the ID of the manifold formed by matrices of different channels in the same image as the local ID estimate of the image.
We suppose that global and local ID should remain invariant as much as possible between the reconstruction by the regularized autoencoder and original inputs.
Our empirical results validate the efficient representation achieved by the encoder of AE-IDC on different downstream tasks.
Our work is limited to the area of DNNs, but it will motivate the future development of other representation-learning algorithms like the probabilistic models and the manifold-learning approaches to exploit the information about the topological structure of the representations' dimensions.

{\small
\bibliographystyle{ieee_fullname}
\bibliography{egbib}
}

\end{document}